\documentclass[letterpaper, 10 pt, conference]{ieeeconf}  

\IEEEoverridecommandlockouts                              

\usepackage{graphics} 
\usepackage{caption}
\usepackage{subcaption}
\usepackage{times} 
\usepackage{amsmath} 
\usepackage{amssymb}  
\usepackage{algorithm}
\usepackage{algpseudocode}

\usepackage{array} 
\usepackage{multirow}
\usepackage{multicol}
\usepackage{nicematrix}
\usepackage{boldline} 
\usepackage{booktabs} 
\usepackage{makecell}

\title{\LARGE \bf
Growing Trees with an Agent: Accelerating RRTs with Learned, Multi-Step Episodic Exploration
}

\author{Xinyu Wu$^{1}$
\thanks{$^{1}$School of Mechanical Engineering and Automation, Harbin Institute of Technology Shenzhen, HIT Campus of University Town of Shenzhen, Shenzhen, China
        {\tt\small xinyu.wuxinyu@outlook.com}}
}

\begin{document}

\maketitle
\thispagestyle{empty}
\pagestyle{empty}

\begin{abstract}
   Classical sampling-based motion planners like the RRTs suffer from inefficiencies, particularly in cluttered or high-dimensional spaces, due to their reliance on undirected, random sampling. This paper introduces the \textbf{Episodic RRT}, a novel hybrid planning framework that replaces the primitive of a random point with a learned, multi-step "exploratory episode" generated by a Deep Reinforcement Learning agent. By making the DRL agent the engine of exploration, ERRT transforms the search process from a diffuse, volumetric expansion into a directed, branch-like growth. This paradigm shift yields key advantages: it counters the curse of dimensionality with focused exploration, minimizes expensive collision checks by proactively proposing locally valid paths, and improves connectivity by generating inherently connected path segments. We demonstrate through extensive empirical evaluation across 2D, 3D, and 6D environments that ERRT and its variants consistently and significantly outperform their classical counterparts without any GPU acceleration. In a challenging 6D robotic arm scenario, ERRT achieves a 98\% success rate compared to 19\% for RRT, is up to 107x faster, reduces collision checks by over 99.6\%, and finds initial paths that are nearly 50\% shorter. Furthermore, its asymptotically optimal variant, ERRT*, demonstrates vastly superior anytime performance, refining solutions to near-optimality up to 29x faster than standard RRT* in 3D environments. Code: {\tt https://xinyuwuu.github.io/Episodic\_RRT/}.
\end{abstract}


\section{INTRODUCTION}

Autonomous navigation in robotics hinges on solving the motion planning problem: computing a feasible, collision-free trajectory through a complex environment \cite{LaValle_2006}.This task is notoriously difficult due to the curse of dimensionality and the high cost of collision checking.

While classical search-based algorithms like A* \cite{Astar1968} are foundational, they scale poorly in high-dimensional continuous spaces. This led to the dominance of sampling-based motion planners (SBMPs) such as the Rapidly-exploring Random Tree (RRT) \cite{lavalle1998rapidly} and Probabilistic Roadmaps (PRM) \cite{PRM1996}, which explore by sampling the space randomly. However, the classical RRT framework is itself inefficient, as its exploration via uncorrelated random points is analogous to a volumetric "mist cloud" that scales poorly and wastes immense computational effort on irrelevant samples \cite{LavalRRTProgress2000} \cite{LBNOIVUD2024}.

\begin{figure}[!htb]
   \centering
   \includegraphics[width=0.7\linewidth]{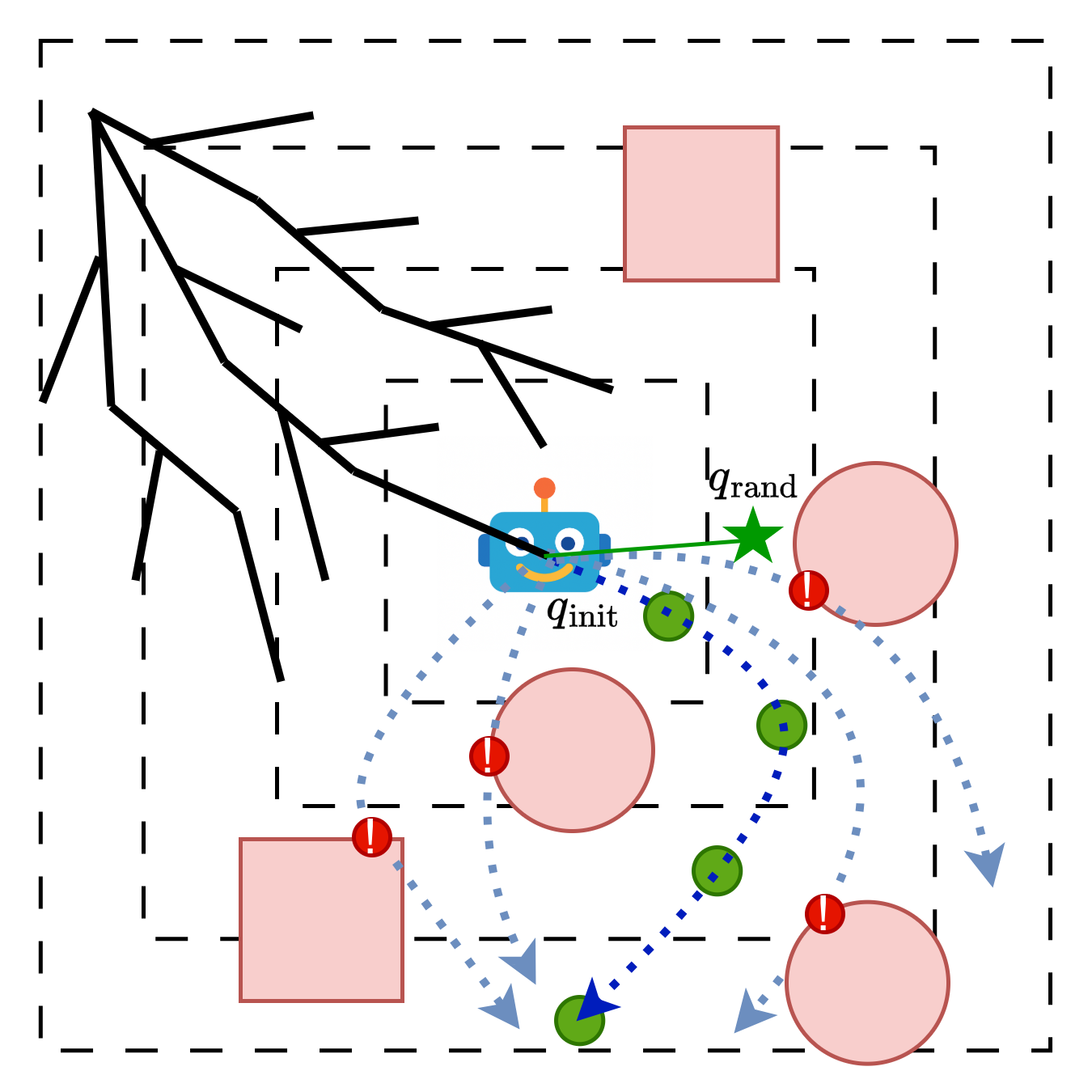}
   \caption{Conceptual illustration of the ERRT framework. Instead of extending the tree with random points, a RL agent proactively generates multi-step "exploration episodes". This transforms the search from a diffuse, random process into a directed, branch-like growth.}
\end{figure}

To overcome these limitations, this paper proposes a paradigm shift with the \textbf{Episodic RRT (ERRT)}, which replaces the primitive of a random point with that of a learned, multi-step "exploratory episode." In our framework, a Deep Reinforcement Learning (DRL) agent acts as the explorer, using a Soft Actor-Critic (SAC) policy \cite{SAC2018} to generate coherent, branch-like path segments. This approach of learned episodic exploration yields three profound and interconnected advantages:

\subsubsection{\textbf{Combating the Curse of Dimensionality}} ERRT replaces volumetric, random exploration with focused, near-linear episodic growth, which is significantly more efficient in high-dimensional spaces where random sampling becomes ineffective.

\subsubsection{\textbf{Minimizing Expensive Computations}} By learning to generate locally valid paths, the DRL agent inverts the costly "sample-first, check-later" paradigm, proactively avoiding collisions and drastically reducing futile computation.

\subsubsection{\textbf{Improving Connectivity and Integration}} The ordered sequence of states in an episode provides inherent via-points, simplifying integration into the tree compared to the difficult and often unsuccessful task of connecting to a single, isolated random point.

The core contribution of this work is the conceptualization and implementation of this novel planning paradigm. We demonstrate through extensive empirical evaluation that ERRT consistently and significantly outperforms classical planners. The remainder of this paper details the related work, presents the ERRT methodology and architecture, provides a comprehensive performance analysis, and concludes with directions for future research.

\section{RELATED WORK}

The quest to improve motion planning efficiency has driven decades of research, primarily along two avenues: refining classical algorithms and integrating machine learning. Our work proposes a novel synthesis of these two fronts.

\subsection{Classical Motion Planning Paradigms}

Foundational to motion planning are search-based algorithms like A* \cite{Astar1968} which find optimal paths on discretized graphs but struggle to scale to high-dimensional continuous spaces. Variants have been developed to address specific challenges, such as the D* family \cite{Dstar1994} \cite{FieldDstar2006} and LPA* \cite{KOENIG200493} for efficient replanning in dynamic environments, Jump Point Search (JPS) \cite{JPS2012} \cite{Harabor_Grastien_2014} for accelerating grid-based searches, and Theta* \cite{Theta*2010} for finding more natural "any-angle" paths.

To avoid explicit discretization, sampling-based motion planners like the Probabilistic Roadmap (PRM) \cite{PRM1996} and Rapidly-exploring Random Tree (RRT) \cite{RRT2001} \cite{lavalle1998rapidly} were introduced. While probabilistically complete, the basic RRT converges to a suboptimal path. RRT* \cite{RRT*PRM*2011} addressed this by introducing a rewiring process to guarantee asymptotic optimality, and RRT-Connect \cite{RRTconnect2000} was developed to accelerate the discovery of an initial solution. Despite their success, these methods are based on sampling uncorrelated points. Subsequent work has focused on biasing this process through heuristics, such as in Informed RRT* \cite{IRRT*2014} or with advanced graph-search techniques like BIT* \cite{BIT*2015} and its successors ABIT* \cite{ABIT*2020} and AIT* \cite{AIT*2020}. However, these advanced methods still operate on a point-wise basis.

\subsection{Learning-Based Motion Planning}

The integration of machine learning has created a spectrum of data-driven planners. At one end, end-to-end generative planners use a learned model to directly produce a complete path in a single pass, as seen in PathRL \cite{PathRL2024} and Neural MP \cite{dalal2024neuralmpgeneralistneural}. While powerful, these methods can lack robustness if the initial prediction is invalid.

A more common approach is to create hybrid planners where a learned model assists a classical framework. In some cases, a neural network replaces a specific algorithmic primitive, acting as a learned subroutine. For example, PRM-RL \cite{PRMRL2018} uses an RL agent to determine if a path exists between two nodes, while RL-RRT \cite{RLRRT2019} uses a policy to execute the steer function. Hierarchical approaches like G2RL \cite{G2RL2020} use a classical global planner and an RL agent for local tracking. Other hybrid methods use the learned model as a weaker form of guidance—a learned heuristic—to make a classical algorithm more efficient. Examples include using a GNN to prioritize edge evaluation \cite{GNNTE2022} or biasing the sampling distribution of an SBMP, as seen in DeepSMP \cite{DeepSMP2018}, CriticalPRM \cite{CriticalPRM2020}, MPNet \cite{MPnet2021}, Neural RRT* \cite{NRRT*2020}, and Neural Informed RRT* \cite{NIRRT*2024}. In these cases, the learned model advises the classical planner but does not change its core "sample-and-connect" logic.

\subsection{Positioning Our Contribution: The RL Agent as an Episodic Explorer}

The ERRT framework introduces a conceptual leap beyond prior art by fundamentally redefining the role of the learned model. Previous hybrid planners typically use a DRL agent in a reactive role: either as a heuristic to bias the sampling of individual points like (e.g., Neural RRT*) or as a subroutine to connect them (e.g., RL-RRT), leaving the core "sample-a-point, connect-the-point" logic intact.

ERRT's primary innovation is to invert this relationship by making the DRL agent the engine of exploration. The DRL agent becomes the proactive explorer, and the fundamental primitive of planning shifts from a single point to a learned, multi-step episode. The growth becomes directed and branch-like, directly countering the curse of dimensionality in a way that simply biasing sample locations cannot. This allows ERRT to retain the robust, incremental structure and theoretical guarantees of RRT while leveraging the power of a learned policy for intelligent, local exploration, achieving a true synthesis of both paradigms, creating a planner that is more than the sum of its parts.

\section{Problem Definition}

In a configuration space $\mathcal{C}$, the general problem of path planning involves finding a path $\sigma: [0,1] \rightarrow \mathcal{C}$ that connects a start configuration $q_{start}$ and a goal configuration $q_{goal}$ while avoiding obstacles. The optimal path planning problem further specifies that the optimal solution path $\sigma^\star$ should be of minimum cost, according to a given cost function $c: \sigma \rightarrow \mathbb{R}$. A key challenge in this problem arises from the division of the configuration space $\mathcal{C}$ into two subspaces: the obstacle space  $\mathcal{C}_{obs}$ and the free space $\mathcal{C}_{free}$. The free space is often implicitly defined as the complement of the obstacle space within the configuration space, represented mathematically as $\mathcal{C}_{free} = \mathcal{C} \setminus \mathcal{C}_{obs}$. The path $\sigma$ is constrained to lie entirely within the free space, represented mathematically as $\sigma(\tau) \in  \mathcal{C}_{free}, \forall \tau \in [0,1]$. The optimal path planning problem is formulated in Equation \eqref{eq:problemDef}.
\begin{equation}
   \begin{aligned}
      \sigma^\star = & \text{arg min } c(\sigma) \\
                     &
      \begin{aligned}
         \text{s.t. } \sigma^\star(0)= & q_{start},\ \sigma^\star(1)= q_{goal}       \\
         \sigma^\star(\tau) \in        & \mathcal{C}_{free},\ \forall \tau \in [0,1]
      \end{aligned}
   \end{aligned}
   \label{eq:problemDef}
\end{equation}

\section{METHOD}

In this section, we first introduce the reinforcement learning process and then introduce the ERRT, ERRT*, and the ERRT-Connect algorithms.

\begin{figure*}[!htbp]
   \centering
   \includegraphics[width=0.9\linewidth]{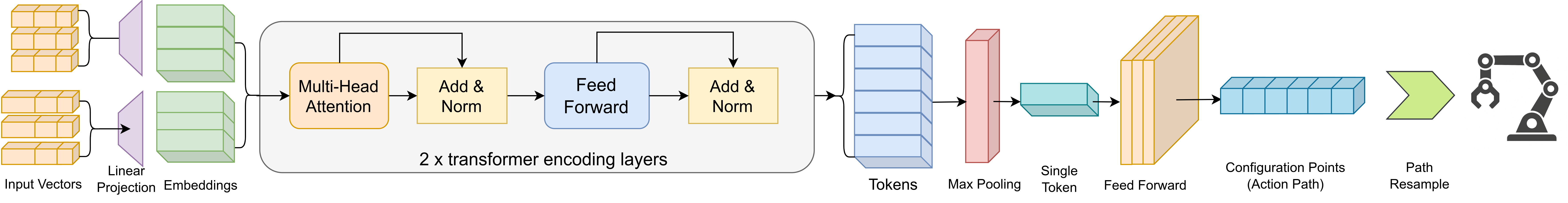}
   \caption{Network Architecture and Data Flow.} \label{fig:net}
\end{figure*}

\subsection{Reinforcement Learning Components}

For Deep Reinforcement Learning, the path planning problem is formulated as a Markov Decision Process (MDP). The MDP is characterized by the tuple $\{ \mathcal{S}, \mathcal{A}, \mathcal{P}, \mathcal{R}, \gamma \}$ which comprises the state space $\mathcal{S}$, action space $\mathcal{A}$, transition probability function $\mathcal{P}$, reward function $\mathcal{R}$ and a discount factor $\gamma \in (0,1)$ that balances the importance of immediate versus future rewards. The core components are defined as follows:

\subsubsection{State Space $\mathcal{S}$} A state $ s_t\in \mathcal{S}$ at time $t$ is a composite vector $s_t = (self_t, goal_t, env_t)$. $self_t$:  The agent's intrinsic state, such as its velocity vector for 2D/3D point-mass planning, or a concatenation of joint positions and velocities for a 6D robotic arm. $goal_t$: The target configuration represented relative to the agent, calculated as $q_{goal} - q_t$. $env_t$: Local environmental context, represented as a variable-length set of feature vectors. Each vector describes an oriented bounding box (OBB) or a bounding sphere within a perception radius $d_o$.

\subsubsection{Action Space $\mathcal{A}$} An action $a_t \in \mathcal{A}$ is conceptualized as a path, formulated as a sequence of $m$ configuration points, $a_t= (q_{t}^1, q_{t}^2, \cdots, q_{t}^m)$. These points collectively define a trajectory segment. The number of points, $m$, is a constant integer, ensuring a fixed action dimensionality of $dim\{\mathcal{A}\} = m \times dim\{\mathcal{C}\}$. Each configuration point within the action path is determined relative to the agent's current configuration.

\subsubsection{Reward Function $\mathcal{R}$} The reward function $\mathcal{R}$ is a weighted sum of several components designed to guide the agent effectively. The immediate reward $r_t$ at time $t$ is calculated as follows: \begin{equation}
   \begin{aligned}
      r_t= & \alpha_1r^{len}_t+\alpha_2r^{smooth}_t+\alpha_3r^{collide}_t+ \\
           & \beta_1r^{reach}_t+\beta_2r^{advance}_t
   \end{aligned}
\end{equation} where $\alpha_1$ to $\alpha_3$ are negative coefficients, while $\beta_1$ and $\beta_2$ are positive coefficients. $r^{len}_t$ is the length of the action path $a_t$. $r^{smooth}_t$ is a measure of path smoothness, quantified based on the dot products of consecutive path segment differentiations and agent velocity. $r^{collide}_t$ is a penalty for collisions. This work uses a refined penalty: \begin{equation}
            r^{collide}_t=\begin{cases}
               1+1/l_{safe} & \text{if collision} \\
               0            & \text{otherwise}
            \end{cases}
         \end{equation}, where $l_{safe}$ is the collision-free path length before impact. This offers a more nuanced signal than a simple binary penalty. $r^{reach}_t$ is a binary reward when the robot reaches the immediate vicinity of target configuration $q_{goal}$. $r^{advance}_t$ is the reduction in distance to the goal achieved by the current step.

\subsubsection{Neural Network Structure} Our network architecture is designed to process a variable number of environmental inputs (i.e., obstacles) and produce a fixed-dimensional action path. The data flow, illustrated in Fig. \ref{fig:net}, begins by encoding the agent's state, the goal, and local obstacle information into a set of input vectors.

Each obstacle is combined with the agent's self-state and goal information to form a unique input vector. Since the raw feature vectors for bounding spheres and oriented bounding boxes result in initial input vectors of differing dimensions, each input vector is first processed by a linear layer. This projects all input vectors into a common, fixed-dimensional embedding space, yielding uniformly sized embeddings. This standardized sequence is then fed into a two-layer Transformer encoder. The Transformer's self-attention mechanism processes the relationships between all obstacles simultaneously, creating a contextually aware representation of the environment.

To distill this information, a 1D max-pooling operation is applied across the feature dimensions of these output tokens from the Transformer into a single, fixed-dimensional vector that holistically represents the surrounding environment. Finally, this vector is passed through a feedforward network that generates the final, fixed-dimensional action path irrespective of the initial number of obstacles. This architecture allows the agent to effectively reason about complex scenes with any number of obstacles.

\begin{algorithm}[!htbp]
   \caption{Reinforcement Learning Training}
   \label{alg:RL_training}
   \begin{algorithmic}[1] 
      \State \textbf{Input:} Initial policy parameters $\theta$; Q-function parameters $\phi_1, \phi_2$; empty replay buffer $\mathcal{D}$; maximum retries per state $max_{re}$.
      \State Initialize target network parameters: $\phi_{\text{targ},1} \gets \phi_1$, $\phi_{\text{targ},2} \gets \phi_2$.
      \Repeat
      \State Observe current state $s$.
      \State \parbox[t]{0.9\linewidth}{Select action $a \sim \pi_\theta(\cdot|s)$ and apply Incremental Bound during the generation of $a$.}
      \State Re-sample action path $a_r \gets \text{PathReSample}(a)$.
      \State Execute $a_r$ in the environment.
      \State Observe next state $s'$, reward $r$, and done signal $d$.
      \State Store the transition $(s, a, r, s', d)$ in replay buffer $\mathcal{D}$.
      \If{$d=\text{true}$}
      \If{collision occurred \textbf{and} $max_{re}$ not reached}
      \State \parbox[t]{0.8\linewidth}{Reset environment to state $s$ to attempt a new action sequence from $s$.}
      \Else
      \State \parbox[t]{0.8\linewidth}{Randomly reset environment state to begin a new episode.}
      \EndIf
      \EndIf
      \If{it is time for a training update}
      \State \parbox[t]{0.8\linewidth}{Randomly sample a batch of transitions $B = \{(s_k, a_k, r_k, s'_k, d_k)\}$ from $\mathcal{D}$.}
      \State \parbox[t]{0.8\linewidth}{Update neural networks using the Soft Actor-Critic algorithm with batch $B$.}
      \EndIf
      \Until{convergence criteria are met.}
   \end{algorithmic}
\end{algorithm}

\subsection{Reinforcement Learning Training}

This subsection outlines the reinforcement learning training methodology, which is founded on the Soft Actor-Critic (SAC) algorithm, augmented with several refinements to better serve the path planning problem. These strategies include 'Incremental Bound', 'Path Re-sample', and 'Concentrated Collecting'. The comprehensive training procedure integrating these components is formally presented in Algorithm \ref{alg:RL_training}.

\subsubsection{\textbf{Incremental Bound}} A linear incremental bound is employed for each configuration point in the action path to enhance the stability of the training process. This bound is defined such that for every component $p^{i,j}$ of a configuration point $q^i$, its value must reside within the interval $(-\frac{i}{m} \cdot bound, \frac{i}{m} \cdot bound$), where $bound$ is a positive constant. This constraint is enforced for all $i \in [1,m]$ configuration points and for every $j \in \{1,2, \cdots dim\}$ dimension of these points. In this context, $bound$ is a positive constant, $m$ is the total number of configuration points in the action path, and $dim$ is the dimensionality of the configuration space. This methodology serves to prevent the configuration points from concentrating in a confined area, while still allowing the action path to execute essential maneuvers, like a U-turn.

\subsubsection{\textbf{Path Re-sample}} The action path $a=\{q^1, q^2, \cdots q^m\}$ generated by the neural network serves as control points for the construction of a cubic B-spline. Subsequently, the arc length along this B-spline is computed, and the curve is re-parameterized to achieve equidistant knot points with respect to this arc length. The re-sampled action path is then represented as a sequence of knot points, $a_{r}=\{q^1_r, q^2_r, \cdots q^n_r\}$ where the arc length between consecutive points, $\text{ARC}(q^i_r, q^{i+1}_r)$, is maintained at a constant value $d_{dense}$. This $d_{dense}$ parameter determines the knot point density of the re-sampled path. This re-sampling methodology facilitates a uniform distribution of knot points, renders the point density independent of the neural network's direct output and contributes to a smoother path.

\subsubsection{\textbf{Concentrated Collecting}} Across different steps within an episode, environmental complexity can vary dramatically. To enhance sample efficiency and expedite the training process, more data should be collected in challenging situations. To facilitate data collection proportional to environmental difficulty, an episode is terminated upon collision, and a new episode is initiated from the state immediately preceding the collision. Given that actions generated by the neural network during training are sampled from a normal distribution, and data collection steps are separated by neural network updates, the policy network can be re-executed on the pre-collision observation to yield a new action and subsequent environmental transition. This process of resetting to the previous step is repeated up to a maximum of $max_{re}$ times, or until a collision-free transition is achieved for that step. Consequently, the volume of data collected at a specific step serves as an approximation of the situational difficulty at that point.

\begin{algorithm}[!htbp]
   \caption{ERRT And It's Variants}
   \label{alg:ERRT}
   \begin{algorithmic}[1] 
      \State \textbf{Inputs:} $q_{start}$, $q_{goal}$.
      \State \textbf{Parameters:} $\pi_\theta$, $L_{\text{max}}$, $\alpha_jump$.
      \State \textbf{Initialize}: $\mathcal{T}.init(q_{start})$; $k \gets 0$; $a_r \gets \text{empty path}$.
      \Repeat
      \If{stepEnd}
      \If{episodeEnd} \label{alg:line:episodeEnd}
      \State $q_{rand} \gets \text{SAMPLE}(\mathcal{C})$. \Comment{Random start}
      \State $q_{init} \gets \text{NEAREST}(q_{rand}, \mathcal{T})$. \Comment{from tree}
      \Else \Comment{Concecutive action path}
      \State $q_{init} \gets \text{last element of } a_r$, $n_{init} \gets 0$.
      \EndIf
      \State $a_r \gets \textbf{GenPath}(\pi_\theta, q_{init}, n_{init})$; $k \gets \text{LEN}(a_r)$.
      \Else \Comment{Dynamic Bisection}
      \State $k \gets \textbf{DynamicBisection}(k, \text{isValid})$.
      \EndIf
      \State Update episodeEnd, stepEnd, jump flags.
      \State \textbf{If} episodeEnd but not jump, \textbf{GOTO} line \ref{alg:line:episodeEnd}.
      \If{jump}
      \State $q_{new} \gets q_{goal}$. \Comment{One step Jump}
      \Else
      \State $q_{new} \gets k$ th element of $a_r$. \Comment{Episodic sample}
      \EndIf
      \State $q_{near} \gets \text{NEAREST}(q_{new}, \mathcal{T})$.
      \State isvalid $\gets$ \textbf{not} $\text{COLLISION}(q_{near}, q_{new})$.
      \If{isvalid} \Comment{Extend tree}
      \State $\mathcal{T}.add\_vertex(q_{new})$.
      \State $\mathcal{T}.add\_edge(q_{near}, q_{new})$.
      \EndIf
      \State ERRT*: Rewire and Prune \textbf{if} isvalid.
      \State ERRT-connect: Connect and Swap \textbf{if} episodeEnd.
      \Until{Termination Condition is Met.}
      \State Return $\mathcal{T}$.
   \end{algorithmic}
\end{algorithm}

\subsection{The Episodic RRT Algorithm}

The ERRT framework marks a significant evolution of the classical RRT by replacing scatter-shot random sampling with a more intelligent, structured exploration strategy. This is achieved through episodic exploration, where a trained RL policy , $\pi_\theta$, generates coherent, multi-step path segments, or "episodes," to grow the tree.The core logic elegantly integrates path generation and validation into a single, cohesive loop, with minor modifications accommodating the ERRT* and ERRT-Connect variants.

The algorithm's main loop orchestrates the exploration process, which can be broken down into four key steps:

\subsubsection{\textbf{Episode Initialization by Randomized Restart}} The algorithm maintains the state of the current exploration episode. An exploration episode concludes when either its length reaches $L_{max}$ or a policy-generated path is substantially obstructed (i.e., Dynamic Bisection converges in the middle of an action step). When a new episode begins, the algorithm re-introduces a randomized restart to ensure probabilistic completeness. It samples a random configuration, $q_{rand}$, and selects the nearest node in the existing tree, $q_{init}$, as the starting point for the new exploration episode. This periodic randomized restart mechanism prevents the policy from becoming trapped in unproductive regions and ensures the entire configuration space remains reachable.

\subsubsection{\textbf{Policy-Guided Path Generation}} From the starting configuration $q_{init}$, the algorithm calls the \textbf{GenPath} helper function, to produce a candidate path segment. This function leverages the power of the trained RL policy, $\pi_\theta$, to generate a deterministic, locally-optimal action path. To enhance robustness and escape potential dead loops, an adaptive noise strategy is employed. If a particular node has been used to start an episode multiple times, Gaussian noise is added to the policy's action, with the variance of this noise growing exponentially with the selection count. This adaptive noise injection is crucial for aggressively pushing the exploration away from well-trodden but potentially suboptimal paths, effectively discouraging the planner from getting stuck. The resulting path is then re-sampled into a uniformly dense sequence of configurations, ensuring consistency for the validation step and decoupling the point density from the neural network's direct output.

\subsubsection{\textbf{Validation \& Extension via Dynamic Bisection}} The candidate path generated by the policy is not immediately added to the tree. Instead, it is efficiently checked for collisions using the \textbf{DynamicBisection} helper function, Algorithm \ref{alg:bisect}. This function employs a logarithmic search (bisection) to quickly find the furthest reachable configuration along the candidate path from the existing tree. This bisection is "dynamic" because it acknowledges the evolving tree structure: if an intermediate point is validated and added to the RRT, the bisection's upper search bound is reset to the length of $a_r$, ensuring a re-evaluation of the entire remaining path segment from the newly augmented tree. This is significantly more efficient than a linear, step-by-step check, as it minimizes the number of expensive collision detection calls.

\subsubsection{\textbf{Updates \& One Step Jump}} After each extension attempt, the algorithm updates the state of the episode. It increments the episode length and tracks the minimum distance to the goal achieved during that episode. If the episode concludes (due to length or collision) and the tree has been guided into the immediate vicinity of the goal (i.e., the minimum distance is less than the threshold $\alpha_{jump}$), a "Jump" is attempted. For this work, $\alpha_{jump}$ approximates the maximum feasible distance for a single policy-guided step; therefore, it is called the "One Step Jump". This mechanism tries to directly connect the nearest node in the tree to the goal configuration by setting the candidate for tree extension $q_{new}$ directly to $q_{goal}$. This is not a goal-biasing heuristic in the classical sense, but rather a crucial final step to overcome the precision limitations of the neural network, which may struggle to land exactly on the goal configuration.

The algorithm iterates through these steps, growing the tree with intelligent, multi-step episodes. The optional components for ERRT* (rewiring the tree after each extension to improve path quality) and ERRT-Connect (attempting to connect the start and goal trees at the end of each episode) are integrated seamlessly into this main loop, allowing the core episodic exploration strategy to benefit all variants of the algorithm.

\begin{algorithm}[!tbp]
   \caption{Dynamic Bisection Method}
   \label{alg:bisect}
   \begin{algorithmic}[1] 
      \Function{DynamicBisection}{$k$, $isvalid$}
      \If{$isvalid$}
      \State $k_{lower} \gets k$. \Comment{Update lower bound}
      \State $k_{upper} \gets \text{LEN}(a_r)+1$. \Comment{Dynamic upper}
      \Else
      \State $k_{upper} \gets k$. \Comment{Update upper bound}
      \EndIf
      \If{$k_{upper}-k_{lower}>1$}
      \State $k \gets$ midpoint of $[k_{upper}, k_{lower}]$. \Comment{Bisection}
      \Else
      \State $k \gets -1$. \Comment{Convergence, end of step}
      \EndIf
      \State Return $k$.
      \EndFunction
   \end{algorithmic}
\end{algorithm}

\section{PERFORMANCE EVALUATION AND COMPARISONS}

This section presents a comprehensive empirical evaluation of the proposed ERRT algorithms. All baseline algorithms (RRT, RRT*, RRT-Connect) were implemented using the Open Motion Planning Library (OMPL)\cite{ompl2012}. The proposed ERRT, ERRT*, and ERRT-Connect variants were developed by integrating our reinforcement learning-based episodic sampling framework with these standard OMPL implementations. All experiments were conducted on an Intel Core i7-11700 @ 2.50GHz processor, without GPU acceleration, to ensure fair comparisons of algorithmic efficiency.

\subsection{Experimental Setup and Metrics}

The performance of the algorithms was assessed across three distinct scenario types, designed to cover a range of dimensionalities and complexities, as visualized in Fig. \ref{fig:envs}:

\begin{figure}[!htb]
   \centering
   \begin{subfigure}[t]{0.31\linewidth}
      \centering
      \includegraphics[width=\linewidth]{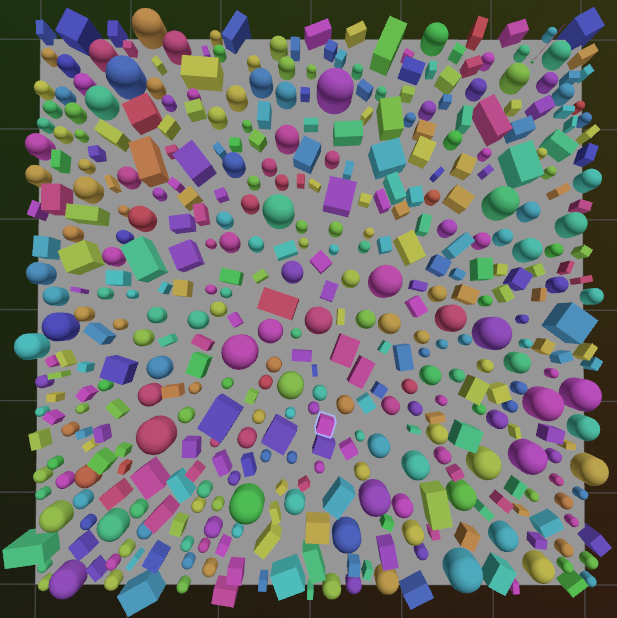}
      \caption{2D}
   \end{subfigure}%
   ~
   \begin{subfigure}[t]{0.31\linewidth}
      \centering
      \includegraphics[width=\linewidth]{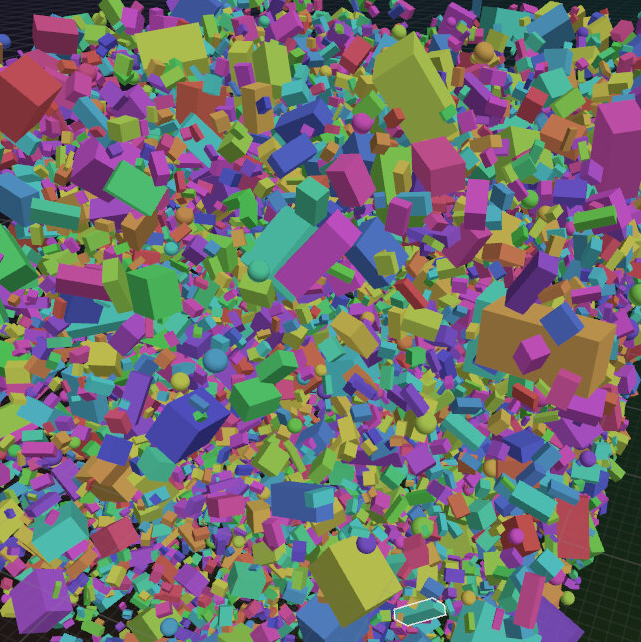}
      \caption{3D}
   \end{subfigure}%
   ~
   \begin{subfigure}[t]{0.31\linewidth}
      \centering
      \includegraphics[width=\linewidth]{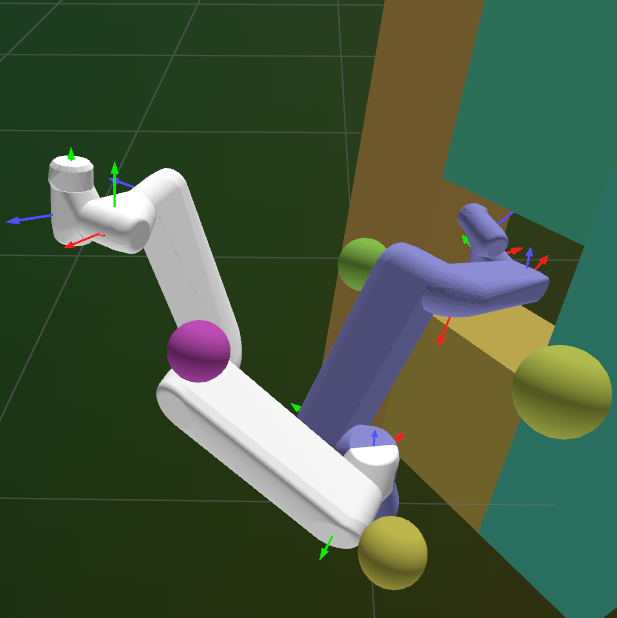}
      \caption{6D}%
   \end{subfigure}
   \caption{Visualization randomly generated environments.} \label{fig:envs}
\end{figure}

2D Scenarios: Cluttered 60x60 m maps with ~400 obstacles whose sizes varied by up to a factor of 30. 3D Scenarios: Dense 60x60x60 m maps with ~17,000 obstacles whose volumes varied by up to a factor of 100. 6D Scenarios: A simulated 6-DOF robotic arm with randomized box and sphere obstacles. To introduce further environmental variability, the box obstacles had a 50$\%$ probability of being configured as thin panels.

For each dimensionality, 100 trials with different random environmental layouts and start/goal pairs were run. Performance was evaluated using the following metrics: Success Rate (\textbf{SUC}), Computation Time (\textbf{TIME}), Collision Checks (\textbf{COL}), and Path Length (\textbf{LEN}). For the 6D robotic arm scenario, \textbf{LEN} is defined as the sum of absolute joint rotations in radians across all six joints.

\subsection{Comparative Analysis of Initial Solutions}

The performance of ERRT variants in finding initial solutions was compared against their OMPL counterparts, with results summarized in Table \ref{tab:compare}. For each dimensionality, 100 distinct random environments were generated. Performance metrics were averaged over these successful trials. Unless otherwise specified, each test run for both baseline and ERRT variants was terminated as soon as the first collision-free path was discovered, without any subsequent path optimization. A default time limit of 1 second per test was imposed; exceptions to this limit for specific challenging baseline runs are noted in the results tables. The analysis, detailed below, reveals several key advantages conferred by the ERRT framework.

\subsubsection{\textbf{Enhanced Robustness}} The ERRT framework demonstrates significantly higher robustness and speed, especially as problem dimensionality increases. In the challenging 6D robotic arm scenario, the standard RRT and RRT* baselines timed out frequently, achieving success rates of only 19\% and 21\% respectively, even with a generous 10-second limit. In stark contrast, their ERRT counterparts were overwhelmingly successful, with ERRT and ERRT* achieving 98\% and 97\% success rates. Even in lower dimensions, baseline RRT* required a much longer time budget to achieve comparable success to ERRT*.

\subsubsection{\textbf{Substantial Speed-Up}} The ERRT methodology provides a considerable reduction in computation time, often by an order of magnitude, and this advantage becomes increasingly pronounced in higher dimensions, growing from a 1.7x speed-up in 2D to 7.9x in 6D for ERRT-Connect versus RRT-Connect. This trend provides strong evidence that the near-linear, branch-like growth of episodic exploration is a fundamentally more effective strategy for scaling than the volumetric expansion of traditional random sampling. This is particularly evident for ERRT*, where the higher quality of samples reduces not only collision checks but also the significant overhead from extensive rewiring operations on poorly chosen samples. In 6D, ERRT* was 82x faster than RRT*.

\subsubsection{\textbf{Drastic Reduction in Collision Checks}} A direct consequence of the more targeted exploration inherent in ERRT is a sharp, order-of-magnitude decrease in the average number of collision checks. As a primary computational bottleneck, this reduction is a critical practical advantage, especially in environments with complex geometries or numerous obstacles. For example, in the 3D environment, ERRT achieved a reduction of over 97\% in collision checks compared to the baseline RRT. The gains were even more significant in the 6D environment, where ERRT reduced collision checks by over 99.6\%. While the neural network inference time for the RL policy adds a small overhead, this is dwarfed by the substantial reduction in collision checks, leading to faster overall planning.

\subsubsection{\textbf{Superior Initial Solution Quality}} A remarkable finding is that even ERRT variants not explicitly designed for optimality (i.e., ERRT and ERRT-Connect) consistently produce significantly shorter, higher-quality initial paths than their respective baselines. For instance, in the 3D scenario, the baseline RRT produced a circuitous path of average length 165.9 m, while ERRT's was 32\% shorter at 112.1 m. This pattern is even more pronounced in 6D, where ERRT's initial path was nearly 50\% shorter than the baseline's (14.61 vs. 28.73 radians). This suggests that the RL policy learns not only to avoid obstacles but also to generate direct, goal-oriented trajectories. This provides a significant advantage over the "random-walk" nature of traditional RRT exploration, which often produces circuitous initial paths.

\begin{table}[tbp]
   \centering
   \setlength{\tabcolsep}{4.5pt} 
   \caption{Initial Solution Comparison}
   \label{tab:compare}
   \begin{NiceTabular}{llcccrc}
      \toprule
      \textbf{DIM}        & \textbf{Methods} & \textbf{SUC$\uparrow$} & \textbf{TIME(s)$\downarrow$} & \textbf{ACC}          & \textbf{COL$\downarrow$} & \textbf{LEN$\downarrow$} \\
      \midrule
      \multirow{6}{*}{2D} & RRT              & 100\%                  & 0.124                        & \multirow{2}{*}{1.7X} & 139965                   & 123.2                    \\
                          & ERRT             & 100\%                  & 0.073                        &                       & 5351                     & 89.96                    \\
      \cmidrule{2-7}
                          & RRT* (2s)        & 96\%                   & 0.910                        & \multirow{2}{*}{9.4X} & 2021510                  & 84.62                    \\
                          & ERRT*            & 100\%                  & 0.096                        &                       & 69947                    & 85.17                    \\
      \cmidrule{2-7}
                          & RRT-con          & 100\%                  & 0.083                        & \multirow{2}{*}{1.7X} & 110864                   & 124.8                    \\
                          & ERRT-con         & 100\%                  & 0.049                        &                       & 4201                     & 91.67                    \\
      \midrule
      \multirow{6}{*}{3D} & RRT              & 94\%                   & 0.459                        & \multirow{2}{*}{2.6X} & 501442                   & 165.9                    \\
                          & ERRT             & 100\%                  & 0.174                        &                       & 14791                    & 112.1                    \\
      \cmidrule{2-7}
                          & RRT* (6s)        & 75\%                   & 2.941                        & \multirow{2}{*}{7.9X} & 3193873                  & 107.2                    \\
                          & ERRT*            & 100\%                  & 0.373                        &                       & 221214                   & 98.70                    \\
      \cmidrule{2-7}
                          & RRT-con          & 100\%                  & 0.164                        & \multirow{2}{*}{2.2X} & 158747                   & 168.9                    \\
                          & ERRT-con         & 100\%                  & 0.074                        &                       & 6077                     & 115.6                    \\
      \midrule
      \multirow{6}{*}{6D} & RRT  (10s)       & 19\%                   & 1.608                        & \multirow{2}{*}{107X} & 87170                    & 28.73                    \\
                          & ERRT             & 98\%                   & 0.015                        &                       & 322                      & 14.61                    \\
      \cmidrule{2-7}
                          & RRT*  (10s)      & 21\%                   & 1.974                        & \multirow{2}{*}{82X}  & 141629                   & 22.49                    \\
                          & ERRT*            & 97\%                   & 0.024                        &                       & 705                      & 13.34                    \\
      \cmidrule{2-7}
                          & RRT-con          & 96\%                   & 0.124                        & \multirow{2}{*}{7.3X} & 4330                     & 26.20                    \\
                          & ERRT-con         & 100\%                  & 0.017                        &                       & 419                      & 16.21                    \\
      \bottomrule
   \end{NiceTabular}
\end{table}

\subsection{Analysis of Convergence to Optimality}

For applications such as unmanned aerial vehicles (UAVs) where path efficiency directly impacts energy consumption, the ability of a planner to rapidly converge to an optimal solution is crucial. We therefore evaluated the anytime performance of ERRT* against the standard RRT* to measure its efficiency in solution refinement in 3D scenario.

\begin{figure}[!htb]
   \centering
   \includegraphics[width=\linewidth]{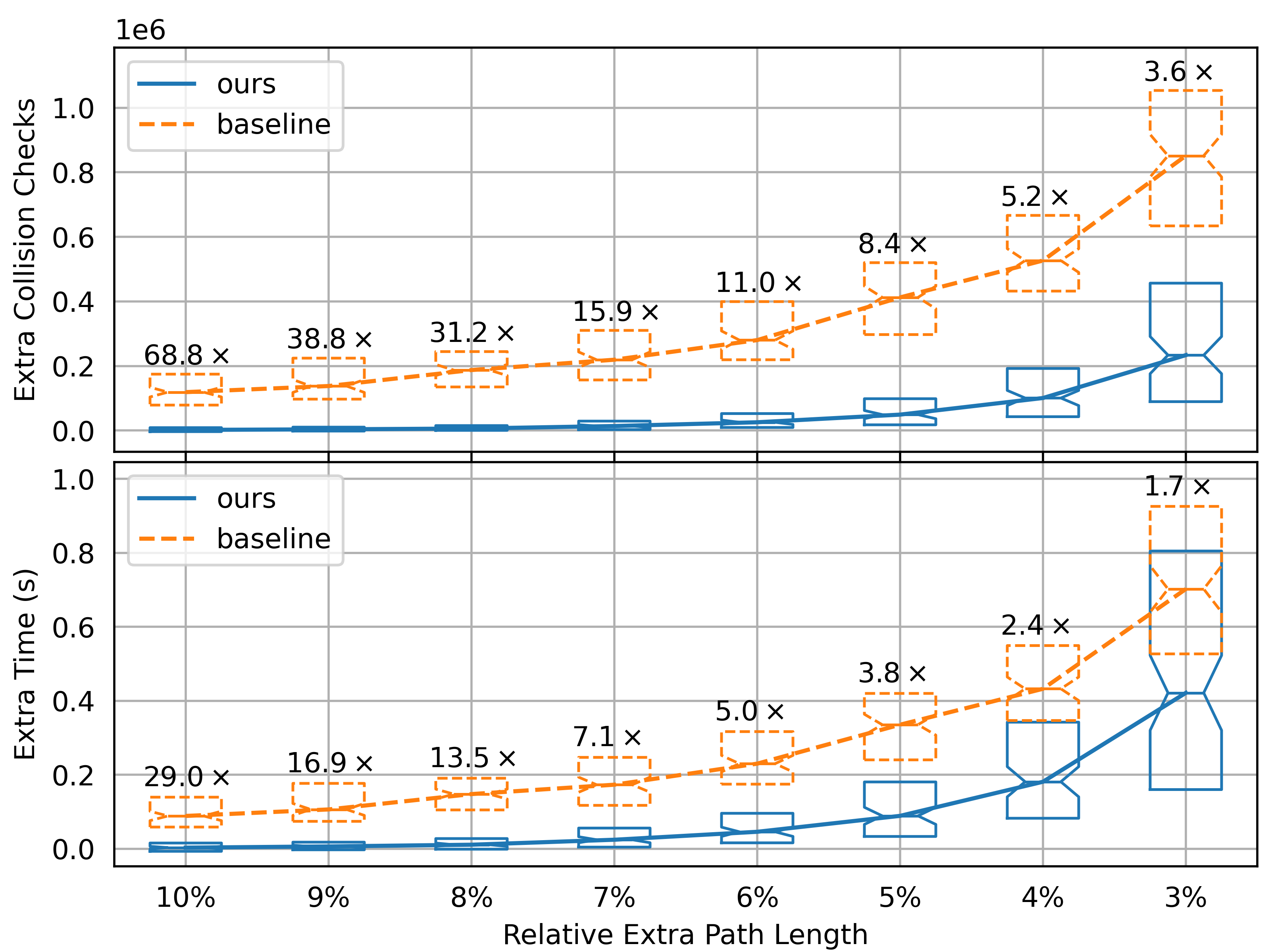}
   \caption{Continuing Optimization Performance.}\label{fig:con-opt}
\end{figure}

To isolate and fairly compare the algorithms' intrinsic convergence capabilities, separate from the initial search efficiency where ERRT is already superior, we measured only the additional computational cost incurred after an initial solution was found. We generated 100 smaller 3D environments (20x20x20 m) and, for each, established an optimal path cost by running RRT* for 10 seconds. We then executed 10 trials per environment for both ERRT* and the baseline RRT*, recording the average extra time and collision checks each algorithm required to refine its initial path to within a series of quality thresholds of the optimal cost.

As shown in Fig. \ref{fig:con-opt}, ERRT* demonstrates vastly superior convergence efficiency. To reach a path just 10\% longer than optimal, ERRT* was already 29.0x faster and required 68.8x fewer additional collision checks than RRT*. This significant performance advantage persists across all quality thresholds. At a 5\% tolerance, ERRT* was 3.8x faster and 8.4x more efficient in collision checks. Even when refining the solution to within 3\% of the optimal path, ERRT* maintained a 1.7x speed advantage and a 3.6x advantage in collision check efficiency.

\subsection{Ablation Study and Parameter Sensitivity}

To validate the design choices of the ERRT framework and assess the contribution of its key components, we conducted a comprehensive ablation study. We systematically disabled or modified core mechanisms and measured the impact on performance relative to the full ERRT implementation in Table \ref{tab:compare}. The results, summarized in Table \ref{tab:ablation}, underscore the criticality of our proposed methods.

\begin{table}[htbp]
   \centering
   \setlength{\tabcolsep}{1.4pt} 
   \caption{Ablation Study}
   \label{tab:ablation}
   \begin{NiceTabular}{ll|cc|cc|cc}
      \toprule

                                     &                  & \multicolumn{2}{c|}{2D} & \multicolumn{2}{c|}{3D} & \multicolumn{2}{c}{6D}                                                \\
      \midrule
      \textbf{Ablation}              & \textbf{Methods} & \textbf{SUC}            & \textbf{TIME}           & \textbf{SUC}           & \textbf{TIME} & \textbf{SUC} & \textbf{TIME} \\
      \midrule
      \multirow{3}{*}{NoBisection}   & ERRT             & -5\%                    & +69 ms                  & -9\%                   & +215 ms       & -1\%         & +5 ms         \\
                                     & ERRT*            & -34\%                   & +597 ms                 & $\times$               & $\times$      & -5\%         & +157 ms       \\
                                     & ERRT-con         & -1\%                    & +33 ms                  & +0\%                   & +73 ms        & -2\%         & +9 ms         \\
      \midrule
      \multirow{3}{*}{\makecell{NoBisection                                                                                                                                         \\ \textbf{\&} \\ DownSample}} & ERRT             & -2\%                    & +32 ms                  & +0\%                   & +22 ms        & -1\%         & +5 ms         \\
                                     & ERRT*            & -7\%                    & +95 ms                  & -34\%                  & +375 ms       & +0\%         & +39 ms        \\
                                     & ERRT-con         & +0\%                    & +0 ms                   & +0\%                   & +16 ms        & +0\%         & +13 ms        \\
      \midrule
      \multirow{3}{*}{DownSample}    & ERRT             & +0\%                    & -2 ms                   & +0\%                   & -13 ms        & -1\%         & -2 ms         \\
                                     & ERRT*            & +0\%                    & -15 ms                  & +0\%                   & -69 ms        & +0\%         & +2 ms         \\
                                     & ERRT-con         & +0\%                    & -21 ms                  & +0\%                   & +9 ms         & +0\%         & +1 ms         \\
      \midrule
      \multirow{3}{*}{HalfStepJump}  & ERRT             & +0\%                    & +2 ms                   & +0\%                   & +12 ms        & -1\%         & -1 ms         \\
                                     & ERRT*            & -1\%                    & +3 ms                   & +0\%                   & -7 ms         & +0\%         & +2 ms         \\
                                     & ERRT-con         & +0\%                    & -22 ms                  & +0\%                   & +5 ms         & +0\%         & -1 ms         \\

      \midrule
      \multirow{3}{*}{NoJump}        & ERRT             & $\times$                & $\times$                & $\times$               & $\times$      & $\times$     & $\times$      \\
                                     & ERRT*            & $\times$                & $\times$                & $\times$               & $\times$      & $\times$     & $\times$      \\
                                     & ERRT-con         & +0\%                    & -26 ms                  & +0\%                   & +1 ms         & +0\%         & +7 ms         \\
      \midrule
      \multirow{3}{*}{2 + $L_{max}$} & ERRT             & +0\%                    & +8 ms                   & +0\%                   & -6 ms         & +0\%         & +3 ms         \\
                                     & ERRT*            & +0\%                    & +7 ms                   & +0\%                   & -67 ms        & +0\%         & -2 ms         \\
                                     & ERRT-con         & +0\%                    & -25 ms                  & +0\%                   & +11 ms        & +0\%         & +1 ms         \\
      \bottomrule
   \end{NiceTabular}
\end{table}

\textbf{Robust and Efficient Path Validation:} The most critical component for ERRT's efficiency is the Dynamic Bisection method for path validation. Disabling this feature (NoBisection) and reverting to a linear check of the densely sampled path proved computationally infeasible. This led to catastrophic failures, with the success rate for ERRT* in 3D dropping to zero, and a prohibitive increase in computation time (e.g., +597 ms for ERRT* in 2D). While reducing the path's sampling density might seem like a solution, doing so without bisection (NoBisection \& Downsample) still incurred a severe performance penalty (e.g., +375 ms for ERRT* in 3D) and a path with a density that is too low risks missing crucial points in cluttered areas. In stark contrast, when Dynamic Bisection was active, changing the sampling density (Downsample) had a negligible impact on performance. This demonstrates that our logarithmic search strategy is essential, as it effectively decouples validation efficiency from path density, eliminating a sensitive trade-off between computational cost and planning reliability.

\textbf{The Role of the One Step Jump:} Our analysis reveals that the "One Step Jump" is not a conventional goal-biasing heuristic but a crucial mechanism for overcoming the precision limitations inherent in neural network outputs. Removing it entirely (NoJump) resulted in a complete failure for the single-tree ERRT and ERRT* methods across all environments. Examination of these failed trials showed the planner consistently guided the tree to within a few centimeters (or a few degrees of rotation in 6D) of the goal but could not bridge the final gap. The bidirectional ERRT-Connect was unaffected, as its dual-tree search provides an alternative connection mechanism. Furthermore, halving the jump distance threshold $\alpha_{jump}$ (HalfStepJump) produced no significant performance degradation. This confirms the mechanism's role as a robust tool for final convergence rather than a sensitive, performance-driving heuristic.

\addtolength{\textheight}{-0.5cm}   

\textbf{Robustness to Hyperparameter Tuning:} Finally, we evaluated the sensitivity of ERRT to its hyperparameters. Increasing the maximum episode length by two steps (2 + $L_{max}$) resulted in only marginal and inconsistent changes to performance. This, combined with the insensitivity to the jump distance threshold $\alpha_{jump}$ and the path re-sample density $d_{dense}$, suggests that the ERRT framework is robust. The core performance gains stem from its architectural design—the episodic exploration and efficient validation—rather than meticulous hyperparameter tuning, highlighting its practical applicability.

\section{Real-World Deployment}

To validate the practical applicability of the ERRT framework, we deployed the planner on a physical UR5e robotic arm for tasks requiring it to maneuver around obstacles. For execution, we first applied the TOPP-RA\cite{toppra2018} algorithm for time-optimal path parameterization and then used a PID controller to track the resulting trajectory. The ERRT planner, using a policy trained entirely in simulation, consistently and rapidly generated smooth, collision-free paths on the physical hardware, demonstrating that the learned policy generalizes effectively to the real world when integrated within the hybrid framework.

\section{conclusion}

This paper's Episodic RRT framework uses a DRL agent to replace inefficient random sampling with learned, obstacle-avoiding exploration. This approach yields dramatic performance gains, including a 107x speed-up and a 99.6\% reduction in collision checks in a 6D environment. Having addressed sampling inefficiency from environmental clutter, the key future direction is to extend this paradigm to also learn complex kinodynamic constraints, such as turning radii and acceleration limits, to tackle the next major challenge in sampling-based planning.










\bibliographystyle{./IEEEtran}
\bibliography{./IEEEabrv,./IEEEexample}

\end{document}